\documentclass[10pt,twocolumn,letterpaper]{article}

\usepackage{iccv}
\usepackage{times}
\usepackage{epsfig}
\usepackage{graphicx}
\usepackage{amsmath}
\usepackage{amssymb}
\usepackage{subcaption}
\usepackage{floatrow}
\usepackage[]{algorithm}
\usepackage[]{algorithmic}
\usepackage{multirow}

\usepackage{tabulary,multirow,overpic,xcolor}
\usepackage{caption,tabularx,booktabs}

\newfloatcommand{capbtabbox}{table}[][\FBwidth]

\usepackage[pagebackref=true,breaklinks=true,letterpaper=true,colorlinks,bookmarks=false]{hyperref}

\iccvfinalcopy % *** Uncomment this line for the final submission

\newcolumntype{x}[1]{>{\centering\arraybackslash}p{#1pt}}

\newcommand{\app}{\raise.17ex\hbox{$\scriptstyle\sim$}}

\def\x{$\times$}
\newcolumntype{x}[1]{>{\centering\arraybackslash}p{#1pt}}

\newlength\savewidth\newcommand\shline{\noalign{\global\savewidth\arrayrulewidth
  \global\arrayrulewidth 1pt}\hline\noalign{\global\arrayrulewidth\savewidth}}

\makeatletter\renewcommand\paragraph{\@startsection{paragraph}{4}{\z@}
  {.5em \@plus1ex \@minus.2ex}{-.5em}{\normalfont\normalsize\bfseries}}\makeatother

\ificcvfinal\fi
\begin{document}
\newcommand{\ours}{FAST-GRU\xspace}
\newcommand{\oursfr}{FASTER\xspace}
\newcommand{\sota}{state-of-the-art\xspace}
\newcommand{\specialcell}[2][c]{\begin{tabular}[#1]{@{}c@{}}#2\end{tabular}}

\title{FASTER Recurrent Networks for Efficient Video Classification}

\author{Linchao Zhu $^\dag$$^\S$\hspace{6mm}
Laura Sevilla-Lara $^\dag$$^*$\hspace{6mm}
Du Tran$^\dag$ \hspace{6mm}
Matt Feiszli$^\dag$ \hspace{6mm}
Yi Yang$^\S$ \hspace{6mm}
Heng Wang$^\dag$ \hspace{6mm}
\\
$^\dag$Facebook AI
\hspace{6mm}
$^\S$University of Technology Sydney
\hspace{6mm}
$^*$University of Edinburgh
}

\maketitle

\begin{abstract}
Typical video classification methods often divide a video into short clips, do inference on each clip independently, then aggregate the clip-level predictions to generate the video-level results. However, processing visually similar clips independently ignores the temporal structure of the video sequence, and  increases the computational cost at inference time. In this paper, we propose a novel framework named \oursfr, \ie, Feature Aggregation for Spatio-TEmporal Redundancy. \oursfr aims to leverage the redundancy between neighboring clips and reduce the computational cost by learning to aggregate the predictions from models of different complexities. The \oursfr framework can integrate high quality representations from expensive models to capture subtle motion information and lightweight representations from cheap models to cover scene changes in the video.
A new recurrent network (\ie, \ours) is designed to aggregate the mixture of different representations. Compared with existing approaches, 
\oursfr can reduce the FLOPs by over $10\times$ while maintaining the state-of-the-art accuracy across popular datasets, such as Kinetics, UCF-101 and HMDB-51.
\end{abstract}

%%%%%%%%% ABSTRACT
\section{Introduction}

\begin{figure}[t]
\centering
\begin{subfigure}[b]{.9\textwidth}
    \centering
    \includegraphics[width=1.\linewidth]{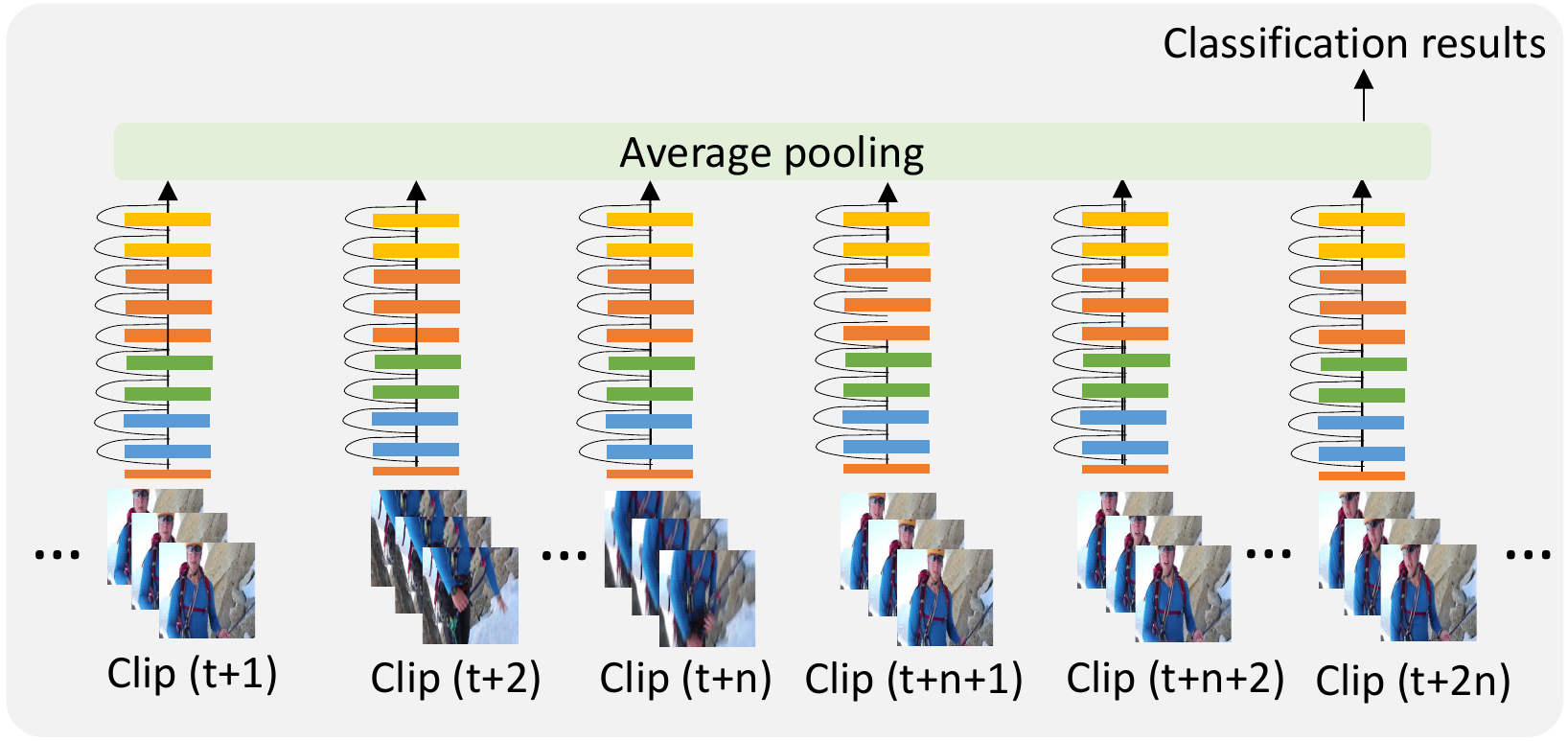}
    \caption{Typical video classification framework for inference.}
    \label{fig:Ng1}
\end{subfigure}
\begin{subfigure}[b]{0.9\textwidth}
    \centering
    \includegraphics[width=1.\linewidth]{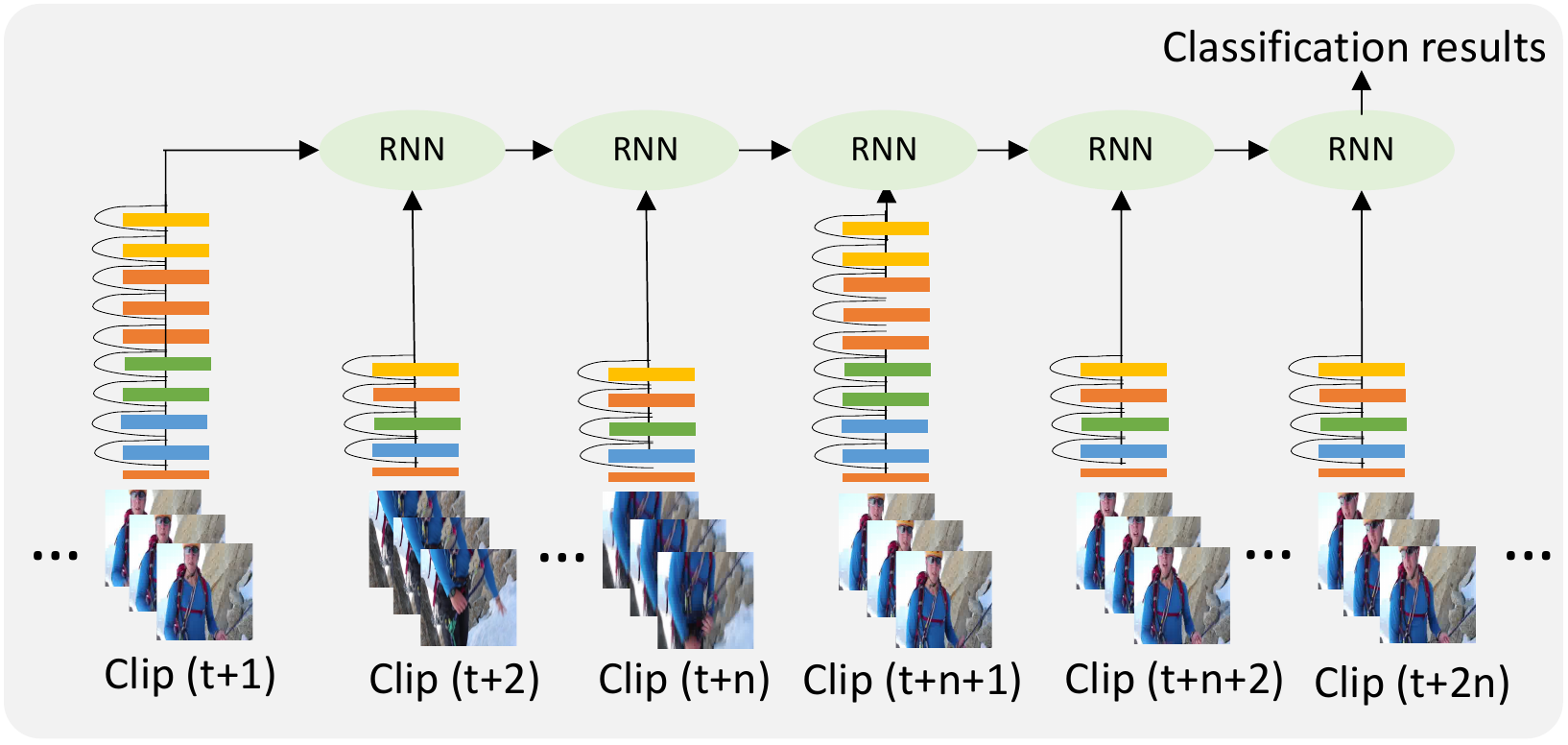}
    \caption{The \oursfr framework can process up to 75\% of the clips using a much cheaper model without losing accuracy.} 
    \label{fig:Ng2}
\end{subfigure}
\caption{(a) Typical video classification framework processes each clip repeatedly with an expensive model; (b) The \oursfr framework exploits the combination of expensive and cheap networks for better accuracy/FLOPs trade-offs.
}
\label{fig:intro}
\end{figure}

Video classification has made tremendous progress since the popularity of deep learning. Though the accuracy of Convolutional Neural Networks (CNNs)~\cite{lecun1998gradient}
on standard video datasets continues to improve, their computational cost is also soaring. For example, on the popular Kinetics dataset~\cite{kay2017kinetics}, the pioneering C3D~\cite{c3d} reported a top-1 accuracy of $63.4\%$ with a single-clip FLOPs of 38.5G. The recent I3D~\cite{carreira2017quo} model improved the accuracy to an impressive $72.1\%$ while its FLOPs also increased to 108G. Non-local networks~\cite{wang2018non} achieved the state-of-the-art accuracy of $77.7\%$ on Kinetics with FLOPs as high as 359G. Moreover, this accuracy is achieved by sampling $30$ clips from each video, which increases the computational cost by a factor of $30$ at test time, as shown in Fig.~\ref{fig:intro} (a). Though modern architectures have achieved incredible accuracy for action recognition, their high computational costs prohibit them from being widely used for real-world applications such as video surveillance, or being deployed on hardware with limited computation power, like mobile devices.

Fig.~\ref{fig:intro} (a) illustrates the typical framework for video classification. Multiple clips are sampled from a video, and  fed to a computationally expensive network to generate predictions for each clip. Clip-level predictions are then aggregated, often by taking the average, to form the final video-level results. To reduce the computational cost, recent efforts have focused on designing more efficient architectures~\cite{howard2017mobilenets} at the clip-level. However, less attention has been paid to the efficiency of the overall framework, including how to aggregate the clip-level predictions over time, since they have a strong temporal structure. 

In this work we address the problem of reducing the computational cost of video classification by focusing on the temporal aggregation stage. In particular, we leverage the observation that video data has a strong temporal structure and is highly redundant in time. We argue that it is computationally inefficient to process many video clips close in time with an expensive video model. 
As adjacent clips are visually similar, computational cost can be saved by processing most of clips with a lightweight network, shown in Fig.~\ref{fig:intro} (b). A Recurrent Neural Network (RNN) is learned to aggregate the representations from both the expensive and lightweight models. We summarize the contributions of our paper in the following.

First, we propose a novel framework for efficient video classification that we call {\em \oursfr} for Feature Aggregation for Spatio-TEmporal Redundancy (Fig.~\ref{fig:intro} (b)). \oursfr explicitly leverages the redundancy of video sequences to reduce FLOPs. Instead of processing every clip with an expensive video model, \oursfr uses a combination of an expensive model that captures the details of the action, and a lightweight model which captures scene changes over time, avoids redundant computation, and provides a global coverage of the entire video at a low cost.
We show that up to 75\% of the clips can be processed with a much cheaper model without losing classification accuracy, maintaining the \sota accuracy with 10$\times$ less FLOPs.

Second, we design a new RNN architecture, namely \ours, dedicated to the problem of learning to aggregate the representations from different clip models. As shown in our experiments, the prevailing average pooling fails to encode predictions from different distributions. On the contrary, \ours is capable of learning integration of representations from multiple models for longer periods of time than popular RNNs, such as GRU~\cite{Cho_GRU} and LSTM~\cite{lstm}. Compared with typical GRU, \ours  keeps the spatio-temporal resolution of the feature maps unchanged, whereas GRU collapses the resolution with global average pooling. \ours also has a channel reduction mechanism for feature compression to reduce the number of parameters and introduce more non-linearity. As a result, the \oursfr framework achieves significantly better accuracy/FLOPs trade-offs for video classification. 
\section{Related Work}
\label{sec:related_work}
Before the deep learning era, hand-crafted features~\cite{wang2011action,wang2013action,laptev2005space,klaser2008spatio,dalal2006human} were widely used for action recognition. In this section, we cover related works that apply deep learning for efficient video classification.

\paragraph{Clip-level Classification.}
Many efforts in video classification focus on designing models to classify short clips. 3D convolutions~\cite{baccouche2011sequential,ji20133d,karpathy2014large,c3d} are widely adopted for learning spatio-temporal information jointly. 
While these methods do improve the accuracy, they also dramatically increases computational cost due to expensive 3D convolutions.
Recent progress includes factorizing 3D convolution into 2D spatial convolution and 1D temporal convolution~\cite{sun2015human,qiu2017learning,tran2018closer,Xie_2018_ECCV,zhou2018mict}, and designing new building blocks to learn global information on top of 3D convolution~\cite{wang2018non,chen2018multi}. Another line of research designs two-stream networks~\cite{simonyan2014two,yue2015beyond,wang2016temporal,feichtenhofer2016convolutional,carreira2017quo,Xie_2018_ECCV} that use both RGB and optical flow as input. Though two-stream networks tend to improve the accuracy, accurate optical flow algorithms are often computationally expensive.

\paragraph{Video-level Aggregation.}
To aggregate the clip-level predictions, average pooling is the most popular approach and works remarkably well in practice.
Besides average pooling, \cite{karpathy2014large} proposed convolutional temporal fusion networks for aggregation.
NetVLAD~\cite{arandjelovic2016netvlad} also has been used in~\cite{girdhar2017actionvlad} for feature pooling on videos.
\cite{yue2015beyond} learned a global video representation by max-pooling over the last convolutional layer.
\cite{wang2016temporal} proposed the TSN network that divides the video into segments, which are averaged for classification.
RNNs are also used for sequential learning in video classification~\cite{icml2015_srivastava,yue2015beyond,li2018videolstm}.
In contrast with existing work,  \ours  aggregates the representations from different models, and aims to leverage spatio-temporal redundancy for efficient video classification.

\paragraph{Towards Efficient Video Classification.} 
\cite{zolfaghari2018eco} proposed the ECO model for online video understanding, where a 3D network is used to aggregate frame-level features.  Recent work~\cite{alwassel2018actiOn,wu2019adaframe,korbar2019scsampler,bhardwaj2019efficient} tried to reduce the computational cost by either reducing the number of frames that need to be processed, or sampling discriminative clips to avoid redundant computation. The \oursfr framework is orthogonal to these approaches. Instead of applying sophisticated sampling strategies, we focus on designing a general framework with a new architecture that can learn to aggregate representations from models of different complexities.

\newcommand{\blocks}[3]{\multirow{3}{*}{\(\left[\begin{array}{c}\text{1$\times$1$^\text{2}$, #2}\\ \text{1$\times$3$^\text{2}$, #2}\\[-.1em] \text{1$\times$1$^\text{2}$, #1}\end{array}\right]\)$\times$#3}
}
\newcommand{\blockt}[4]{\multirow{4}{*}{\(\left[\begin{array}{c}\text{{1$\times$1$\times$1}, #2}\\ \text{1$\times$3$\times$3, #3}\\ \text{3$\times$1$\times$1, #2}\\ \text{1$\times$1$\times$1, #1}\end{array}\right]\)$\times$#4}
}
\newcommand{\blockc}[3]{\multirow{4}{*}{\(\left[\begin{array}{c}\text{{1$\times$1$\times$1}, #2}\\ \text{1$\times$3$\times$3, #2}\\ \text{1$\times$1$\times$1, #1}\end{array}\right]\)$\times$#3}
}
\newcommand{\outsizes}[7]{\multirow{#7}{*}{\(\begin{array}{c} \text{\emph{R2D}}: \text{#1$\times$#2$\times$#2}\\ [.1em] \text{\emph{R(2+1)D}}: \text{#4$\times$#5$\times$#5}\end{array}\)}
}

\section{Learning to Aggregate}
\label{sec:architecture}
As shown in Fig.~\ref{fig:intro} (b), \oursfr is a general framework to aggregate expensive and lightweight representations from different clips. In this section ,we investigate how to learn the temporal structure of clips, aggregate diverse representations to reduce computational cost. We first formulate the problem of learning to aggregate, and introduce the \ours architecture for aggregation. We then describe other aggregation operators used for comparison and the backbones for extracting the expensive and lightweight representations.

\subsection{Problem Formulation}

Given a sequence of $N$ clips from a video, we denote their feature representations as $\mathbf{x}_{t}$, where $t \in [0 \dots N-1]$. The problem of learning to aggregate can be formulated as:

\begin{equation}\begin{aligned}
    \mathbf{o}_{t} = f(\mathbf{o}_{t-1}, \mathbf{x}_{t}), t \in [1 \dots N-1],
    \label{eq:recurrent}
\end{aligned}\end{equation}
where $\mathbf{o}_{t-1}$ encodes the historical information before the current clip $\mathbf{x}_{t}$, and $\mathbf{o}_{0}$ equals to $\mathbf{x}_{0}$.  Note that Eq.~\ref{eq:recurrent} is exactly a recurrent neural network. All previous features $\mathbf{x}_{0}$, $\mathbf{x}_{1}$, \ldots, $\mathbf{x}_{t}$ are recursively encoded into $\mathbf{o}_{t}$. The \oursfr  framework is suitable for online applications, which requires to provide classification results at any time.

Note also \oursfr does not make any assumption about the form of $\mathbf{x}_{t}$. In this paper, we use the feature map from the last convolutional layer of the clip models, which is a tensor of shape $l \times h \times w \times c$, as shown in Fig.~\ref{fig:architecture}. $l$ is the temporal length of $\mathbf{x}_{t}$, whereas $ h \times w$ is the spatial resolution and $c$ is the number of channels. At any time $t$, $\mathbf{x}_{t}$ can come from either the expensive or lightweight model.

Once all the features are aggregated, we apply global average pooling to $\mathbf{o}_{N-1} $, then feed it to a fully-connected layer, which is trained with SoftMax loss to generate classification scores.

\subsection{\ours}
\begin{figure}[t]
\centering
\begin{subfigure}[b]{.63\textwidth}
    \centering
    \includegraphics[width=1.\linewidth]{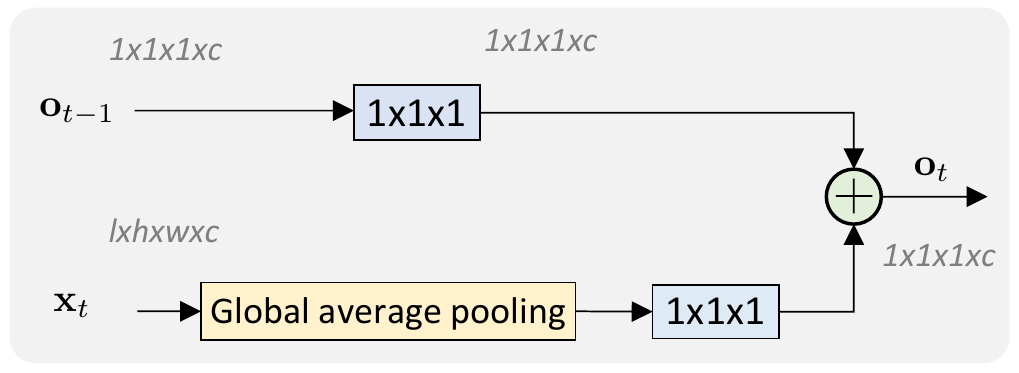}
    \caption{Concat}
    \label{fig:concat_framework}
\end{subfigure}
\begin{subfigure}[b]{0.98\textwidth}
    \centering
    \includegraphics[width=1.\linewidth]{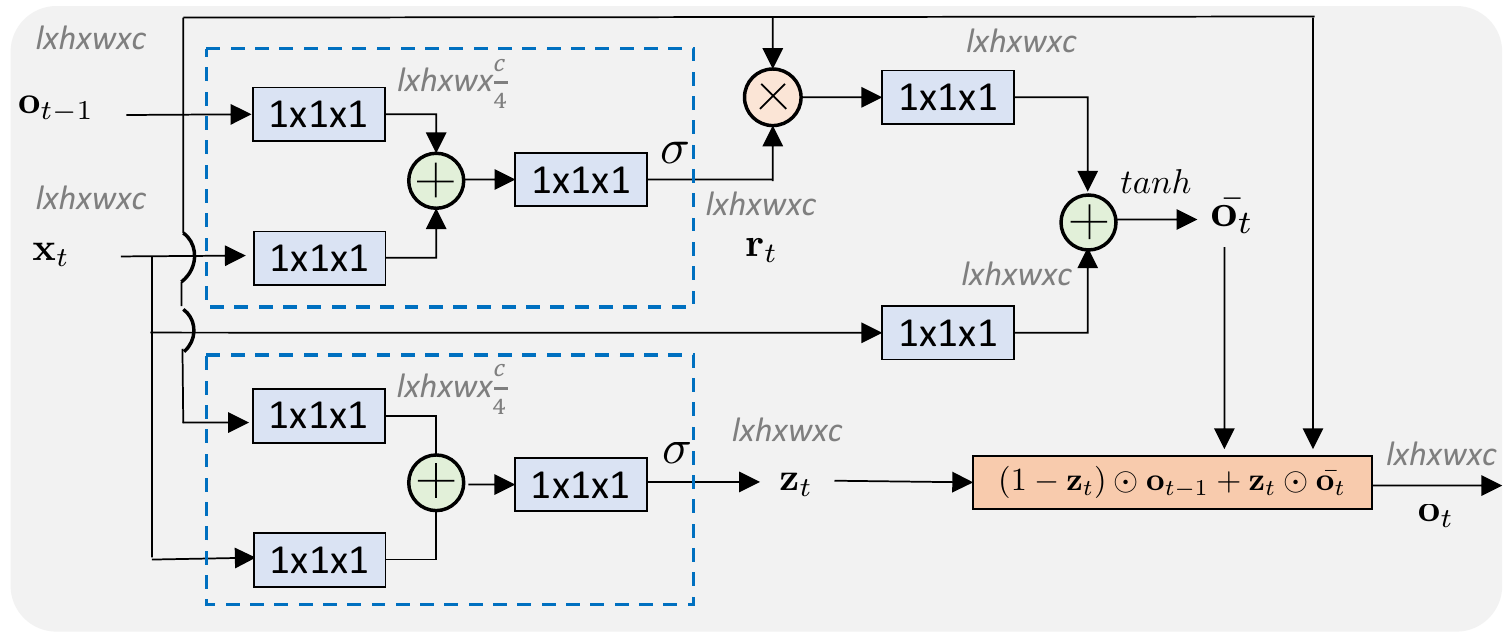}
    \caption{\ours}
    \label{fig:gru_re_framework}
\end{subfigure}
\caption{
(a) The ``Concat" baseline simply concatenates $\mathbf{x}_{t}$ with $\mathbf{o}_{t-1}$; 
(b) Unlike typical GRU collapses the resolution of $\mathbf{x}_{t}$ with global average pooling, \ours keeps the resolution of $\mathbf{x}_{t}$ to learn spatio-temporal information. 
We also introduce bottleneck structures (in dash rectangles) in the gate functions to reduce the number of parameters and increase non-linearity.}
\label{fig:architecture}
\end{figure}

To implement the aggregation function in Eq.~\ref{eq:recurrent}, we propose the \ours architecture shown in Fig.~\ref{fig:architecture} (b). While \ours is based on GRU, we highlight their differences in the following.
\paragraph{3D Convolutional RNN.}
RNNs are usually used to model a sequence of inputs represented as 1D feature vectors.  These feature vectors can be extracted by applying global average pooling on convolutional feature maps.
While these representations are very compact, they do not contain any spatio-temporal information due to pooling. The \oursfr framework requires the RNN to handle representations generated by different models. Thus keeping the spatio-temporal resolution of the feature maps may lead to better performance for aggregation. We empirically show that incorporating  spatio-temporal information is beneficial for long sequence modeling, which has been largely ignored in recent video pooling methods based on RNNs.

Instead of applying a fully-connected layer to the pooled 1D features, we use a 3D $1\times1\times1$ convolution and keep the original resolution of the feature maps in  state transitions, as shown in Fig.~\ref{fig:architecture} (b).  The matrix multiplication in gating functions is also replaced with 3D $1\times1\times1$ convolutions, which enable feature gating in each spatio-temporal location.

\paragraph{Bottleneck Gating.}
In RNNs, the importance of the current input $\mathbf{x}_{t}$ is decided by a subnetwork conditioned on the historical information encoded in $\mathbf{o}_{t-1}$. The subnetwork learns to combine the state  $\mathbf{o}_{t-1}$  and input $\mathbf{x}_{t}$ with different gating mechanisms. Different gates are designed for different functionality for sequential modeling in RNNs. Here we first briefly describe the original GRU architecture, then introduce the bottleneck gating mechanism we proposed. 

In GRU, the subnetwork first concatenates $\mathbf{x}_t$ and $\mathbf{o}_{t-1}$, then project them to a vector of the same dimension, \ie, the number of channels $c$.
There are two gates in GRU, \ie, the read gate $\mathbf{r}$ and update gate $\mathbf{z}$, which are defined as
\begin{equation}\begin{aligned}
    \mathbf{r}_t &=\sigma(\mathbf{G}_{rx}\mathbf{x}_t + {\mathbf{G}_{ro}\mathbf{o}_{t-1}}), \\
    \mathbf{z}_t &=\sigma(\textstyle{\mathbf{G}_{zx}\mathbf{x}_t + {\mathbf{G}_{zo}\mathbf{o}_{t-1}}}), \\
\end{aligned}\end{equation}
where $\mathbf{G}_*$ are  $c\times c$ matrices, and $\sigma$ is the sigmoid function.
After the calculation of $\mathbf{r}_t$ and $\mathbf{z}_t$,
The output of GRU is
\begin{equation}\begin{aligned}
    \bar{\mathbf{o}_t} &= \textstyle{\tanh(\mathbf{V}_{\bar{o}} \mathbf{x}_t + {\mathbf{V}_{\bar{o}} (\mathbf{r}_t \odot \mathbf{o}_{t-1})})}, \\
    \mathbf{o}_t &= \textstyle{(1 - \mathbf{z}_t) \odot \mathbf{o}_{t-1} + \mathbf{z}_t \odot \bar{\mathbf{o}_{t}}}, %, \ \ \ \text{if}\ (t\ \text{MOD}\ T_i) = 0 \\
    %\bar{\mathbf{o}_t} &= \textstyle{\tanh(\mathbf{V}_{\bar{o}} * \mathbf{x}_t + {\mathbf{V}_{\bar{o}} * (\mathbf{r}_t \odot \mathbf{o}_{t-1})})}, \\
    %\mathbf{o}_t &= \textstyle{(1 - \mathbf{z}_t) \odot \mathbf{o}_{t-1} + \mathbf{z}_t \odot \bar{\mathbf{o}_{t}}}, %, \ \ \ \text{if}\ (t\ \text{MOD}\ T_i) = 0 \\
\end{aligned}\end{equation}
where $\mathbf{V}_*$ are $c\times c$ matrices, and $\odot$ represents element-wise multiplication. Both $\mathbf{G}_*$ and $\mathbf{V}_*$ are learnable parameters.

Inspired by ResNet~\cite{he2016deep}, we introduce additional bottleneck structures to increases the expressiveness of the subnetwork in \ours, shown in the dash rectangles of Fig.~\ref{fig:architecture} (b). In the bottleneck layer, $\mathbf{x}_t$ and $\mathbf{o}_{t-1}$ are projected to a lower dimensional feature space after concatenation. This compresses the number of channels to reduce parameters and introduces more non-linearity. The compressed features are then recovered to the original dimension $c$ with another projection. In practice, channel compression is implemented by $1\times 1\times 1$ convolution to reduce the dimension by a factor of $r$. After that, there is an ReLU followed by $1\times 1\times 1$ convolution to recover the dimensionality. The read and update gates of \ours are defined as:
\begin{equation}\begin{aligned}
    \mathbf{r'}_t &= \text{ReLU}(\mathbf{U}_{rx} * \mathbf{x}_t + {\mathbf{U}_{ro} * \mathbf{o}_{t-1}}), \\
    \mathbf{z'}_t &= \text{ReLU}(\textstyle{\mathbf{U}_{zx} * \mathbf{x}_t + {\mathbf{U}_{zo} * \mathbf{o}_{t-1}}}), \\
    \mathbf{r}_t &= \textstyle{\sigma(\mathbf{W}_{r'} * \mathbf{r'})}, \\
    \mathbf{z}_t &= \textstyle{\sigma(\mathbf{W}_{z'} * \mathbf{z'})}, \\
\end{aligned}\end{equation}
where $\mathbf{U}_*$ are $1\times 1\times 1$ convolutions to compress the number of channels by a factor of $r$, and $\mathbf{W}_*$ are $1\times 1\times 1$ convolutions to recover the dimensionality.
Our gates are generated by taking the jointly compressed features $\mathbf{r}_t'$ and $\mathbf{z}_t'$, which enables more powerful gating for feature aggregation. We empirically set $r$ to be $4$, which gives good results in our experiments.

\subsection{Other Aggregation Operator Variants}
Besides \ours, we also consider other baselines for the learning-to-aggregate problem for a comprehensive comparison. We detail each baseline in the following. 

{\bf {\em Average pooling}} is the most popular approach for aggregation. Despite its simplicity,  it  performs remarkably well in practice. In the context of our \oursfr framework, we just average the classification scores from each clip regardless of whether the scores are generated by the expensive or  lightweight model.
The {\bf {\em concat}} baseline is illustrated in Fig.~\ref{fig:concat_framework}, where  $\mathbf{x}_{t}$ and $\mathbf{o}_{t-1}$ are concatenated together, and 
then projected to a joint feature space. It is defined as: 
\begin{equation}\begin{aligned}
    f(\mathbf{o}_{t-1}, \mathbf{x}_{t}) = \text{ReLU}(\text{BN}(\mathbf{W}\mathbf{o}_{t-1} + \mathbf{U}\mathbf{x}_{t})),
    \label{eq:concat_arch}
\end{aligned}\end{equation}
where $\mathbf{W}$ and $\mathbf{U}$ are learnable parameters. % shared across time steps. 
Batch normalization~\cite{ioffe2015batch} and ReLU are also applied in the {\bf {\em concat}} baseline to further improve the performance.

Popular RNNs such as LSTM~\cite{lstm} and GRU~\cite{Cho_GRU} are the most related baselines.  These are the go-to methods for tasks that involve sequential modeling~\cite{sutskever2014sequence}, such as language or speech. For the {\bf {\em LSTM}} baseline, we use the variant that consists of three gates and an additional cell between time steps~\cite{graves2013generating}. LSTM has a forget gate that can reset the history for the current input $\mathbf{x}_{t}$. For video classification, this could be useful in the case of different camera shots or unrelated actions. 
The {\bf {\em GRU}} baseline we used has two gates, \ie, the read and update gate as we described in the previous section.
The state transition function uses the update gate to perform a weighted average of the historical state $\mathbf{o}_{t-1}$ and the current input $\mathbf{x}_{t}$. 

\subsection{Clip-level Backbones}
\oursfr is a general framework and does not make any assumption about the underlying clip-level backbones. We can potentially choose any popular networks, such as I3D~\cite{carreira2017quo}, R(2+1)D~\cite{tran2018closer} or non-local network~\cite{wang2018non}. We implement \oursfr based on  R(2+1)D because it is one of the \sota methods, and its Github repository\footnote{https://github.com/facebookresearch/VMZ} provides a family of networks that can be used as the expensive and lightweight models. In particular, we choose R(2+1)D with 50 layers as the expensive model, and R2D with 26 layers as the lightweight model, as detailed in Table~\ref{tab:arch}.

Comparing with the original R(2+1)D, we make two changes to further improve its performance. First, we replace convolutional blocks with bottleneck blocks, which have been widely used in the family of ResNet architectures and shown to both reduce computational cost and improve accuracy. Second, we insert a max-pooling layer after $\mathbf{conv}_1$, which enables R(2+1)D to support a spatial resolution of $224\times224$ without significantly increasing its FLOPs.  

For the lightweight R2D model, bottleneck layers are used in the same way as R(2+1)D. To reduce the FLOPs of R2D, a temporal stride of 8 is used in $\mathbf{conv}_1$, which effectively reduces the temporal length of the clip by a factor of 8. Unlike R(2+1)D, R2D only has 26 layers to further reduce the computational cost.

To integrate R(2+1)D-50 and R2D-26 in \oursfr, we simply use the two backbones as feature extractors. In Table~\ref{tab:arch}, the output from res$_5$ generates a feature map of size $\frac{L}{8} \times 7 \times 7 \times 2048$, which is used as input $\mathbf{x}_{t}$ to the RNNs.

\begin{table}[t]
\centering
\resizebox{\columnwidth}{!}{
    \begin{tabular}{c|c|c|c}
        \hline
        Layers & R2D-26      &  R(2+1)D-50  & Output sizes $L$\x$H$\x$W$ \\
        \shline
        \multirow{2}{*}{conv$_1$} & \multicolumn{1}{c|}{{8\x7\x7}, {64}}  & \multicolumn{1}{c|}{{1\x7\x7}, {45}, stride 1,2,2}  &  \outsizes{{$\frac{L}{8}$}}{112}{64}{{$L$}}{112}{8}{2}   \\
                               & stride 8, 2, 2                          &  3\x1\x1, 64, stride 1,1,1  & \\
        \hline
        \multirow{2}{*}{pool$_1$}  &  \multicolumn{2}{c|}{1\x3\x3 max}  & \outsizes{{$\frac{L}{8}$}}{56}{64}{{$L$}}{56}{8}{2} \\
         & \multicolumn{2}{c|}{stride 1, 2, 2}  &    \\
        \hline
        \multirow{4}{*}{res$_2$} & \blockc{{256}}{{64}}{2} & \blockt{{256}}{{64}}{144}{3}  & \outsizes{{$\frac{L}{8}$}}{56}{256}{{$L$}}{56}{32}{4}  \\
        &  & & \\
        &  & & \\
        &  & & \\
        \hline
        \multirow{4}{*}{res$_3$} & \blockc{{512}}{{128}}{2}  &  \blockt{{512}}{{128}}{288}{4}  & \outsizes{{$\frac{L}{8}$}}{28}{512}{{$\frac{L}{2}$}}{28}{64}{4}  \\
        &  & & \\
        &  & & \\
        &  & & \\
        \hline
        \multirow{4}{*}{res$_4$} & \blockc{{1024}}{{256}}{2}  & \blockt{{1024}}{{256}}{576}{6}  &  \outsizes{{$\frac{L}{8}$}}{14}{1024}{$\frac{L}{4}$}{14}{128}{4}  \\
        &  & & \\
        &  & & \\
        &  & & \\
        \hline
        \multirow{4}{*}{res$_5$} & \blockc{{2048}}{{512}}{2} & \blockt{{2048}}{{512}}{1152}{3}   &   \outsizes{{$\frac{L}{8}$}}{7}{2048}{$\frac{L}{8}$}{7}{256}{4} \\
        &  & & \\
        &  & & \\
        &  & & \\
        \hline
        \multicolumn{3}{c|}{global average pooling, fc}  &  \\
        \hline
    \end{tabular}}
    \caption{{\bf Clip-level backbones for extracting the expensive and lightweight representations.} The FLOPs of R(2+1)D-50 is about 10$\times$ of R2D-26.}
\label{tab:arch}
\end{table}

\section{Experimental Setups}
In this section, we describe the experimental setups, \ie, the datasets, training and test protocols for both the clip-level backbones and \oursfr framework.

\paragraph{Datasets.}
We choose the Kinetics~\cite{kay2017kinetics} dataset as the major testbed for \oursfr. Kinetics is among the most popular datasets for video classification. 
To simplify, all reported results on Kinetics are trained from scratch, without pre-training on other datasets (\eg, Sports1M or ImageNet).
Kinetics has 400 action classes and about 240K training videos. We report top-1 accuracy on the validation set as  labels on the testing set is not public available.
We also report results on UCF-101~\cite{soomro2012ucf101} and HMDB-51~\cite{kuehne2011hmdb}. These datasets are much smaller, thus we use Kinetics for pre-training and report  mean accuracy on three testing splits.

\begin{table}[t]
\footnotesize
\resizebox{\columnwidth}{!}{
    \centering
\begin{tabular}{c|c|c|c}
\hline
    Model & Input           & Video$@$1  & GFLOPs\x Clips \\
\shline
\multirow{3}{*}{R2D-26} & 8-frames     &     63.3       &  3.2\x32   \\
                        & 16-frames      &    65.5        &     6.0\x16   \\
                        & 32-frames        &    66.7        &     12.7\x8    \\
\hline
\multirow{3}{*}{R(2+1)D-50} & 8-frames      &   70.1         &   30.0\x32       \\
                            & 16-frames      &   72.3         &   60.0\x16      \\
                            & 32-frames     &   74.5         &   119.9\x8      \\
\hline
\end{tabular}}
\caption{{\bf Accuracy of clip-level backbones on Kinetics.} 
The significant performance gap between the two backbones offers a great opportunity to explore better accuracy/FLOPs trade-offs when combining them.  }
\label{table:backbone}
\end{table}

Due to limited GPU memory, we train the \oursfr framework in a two-stage process. First, two clip-level backbones are trained by themselves separately. In the second stage, two backbones are fixed and we train different RNNs using the feature maps extracted by the backbones.

\paragraph{Setups for clip-level backbones.}
We mostly follow the procedure in \cite{tran2018closer} to train the clip-level backbones except two changes. First, we scale the input video whose shorter
side is randomly sampled in [256, 320] pixels, following~\cite{wang2018non,simonyan2014very}. Second, we adopt the cosine learning rate schedule~\cite{loshchilov2016sgdr}. During training, we randomly sample $L$ consecutive frames from a given video. For testing, we uniformly sample $N$ clips to cover the whole video, and average pool the classification scores of each clip, as shown in Table~\ref{table:backbone}. We fix the total number of frames processed to be 256, \ie, $N \times L = 256$. As the average length of Kinetics videos is about 10 seconds in 30 FPS, 256 frames give us enough coverage over the whole video. We only use RGB frames as input as computing optical flow can be very expensive.

\paragraph{Setups for learning to aggregate.}
To train different RNNs for learning to aggregate, we randomly sample $N$ clips, and each clip has $L$ consecutive frames. As mentioned above, the clip-level backbones are fixed  in the second stage due to limited GPU memory. The initial state $\mathbf{o}_0$ is set to be the features from the first clip $\mathbf{x}_{0}$. The training procedure is similar to the first stage except we train much fewer epochs. For testing, the $N$ clips are uniformly sampled to cover the whole video.

\begin{table}[t]
\centering
\resizebox{0.9\columnwidth}{!}{
\centering
\begin{tabular}{c|ccccc}
\hline
    \multirow{2}{*}{Model}               & \multicolumn{5}{c}{Number of Clips} \\
             & 2 & 4 & 8 & 16 & 32  \\
        \shline
            Avg. pool & 60.7 & 63.9 & 64.2  & 64.0  & 63.8 \\
            Concat & 61.6 & 65.9 & 66.3 & 62.3 & 58.2 \\
            LSTM & 61.3  & 66.2 & 66.3 & 66.3 & 64.7 \\
            GRU &  61.1 & 65.6 & 66.5 & 66.1  & 65.8 \\
            \ours & 61.1 & 65.6 & \textbf{67.2} & \textbf{67.4} & \textbf{67.2}  \\ \hline
\end{tabular}}
\caption{{\bf Comparison of different  architectures for aggregation on Kinetics.} The clip length $L$ is 8. For all the methods, only  the first clip is processed by the expensive model, and the remaining clips are processed by the lightweight model.} 
\label{tab:rnn_variations}
\end{table}

\section{Experimental Results}
In this section, we demonstrate the advantage of \oursfr and discuss accuracy/FLOPs trade-offs in different settings  on Kinetics. We also compare \oursfr with the state of the art on Kinetics, UCF-101, and HMDB-51.

\paragraph{Accuracy of the backbone architectures.} 
Table.~\ref{table:backbone} summarizes the top-1 accuracy of the two clip-level backbones on Kinetics. We consider the clip length $L$ to be 8, 16 or 32 frames, and sample 32, 16 or 8 clips, respectively. This is to ensure that we always process the same number of frames under different settings. Note that the GFLOPs of R(2+1)D-50 is about 10$\times$ of R2D-26, and the accuracy of R(2+1)D-50 is about $7\%$ better than R2D-26. The significant performance gap between the two backbones gives us the opportunity to explore different trade-offs in the following experiments. For the same backbone, longer clip length $L$ often leads to better accuracy. But we can not train R(2+1)D-50 with $L=64$ frames due to GPU memory limit.

\paragraph{\ours \vs. other RNN variants.} 
We compare the accuracy of different architectures for the learning-to-aggregate problem  across a wide range of number of clips ($N=\{2, 4, 8, 16, 32\}$). The first clip is always processed with the expensive model and the following $N-1$ clips are processed with the cheap model. We use clip length $L=8$ frames. This allows us to observe the behavior of different architectures in shorter and longer sequences. The results are shown in Table~\ref{tab:rnn_variations}. For ``Avg. pool'', we simply average the classification scores of all the clips. The performance of ``Avg. pool'' saturates as more cheap clips are added. Less accurate predictions dominate the final score, since the accuracy of 16 clips (\ie, 64.0\%) is quite similar to the accuracy of 32 clips (\ie, 63.8\%). This suggests that the expensive representations from the first clip are not fully utilized.

\begin{figure}[t]
\includegraphics[width=1.\linewidth]{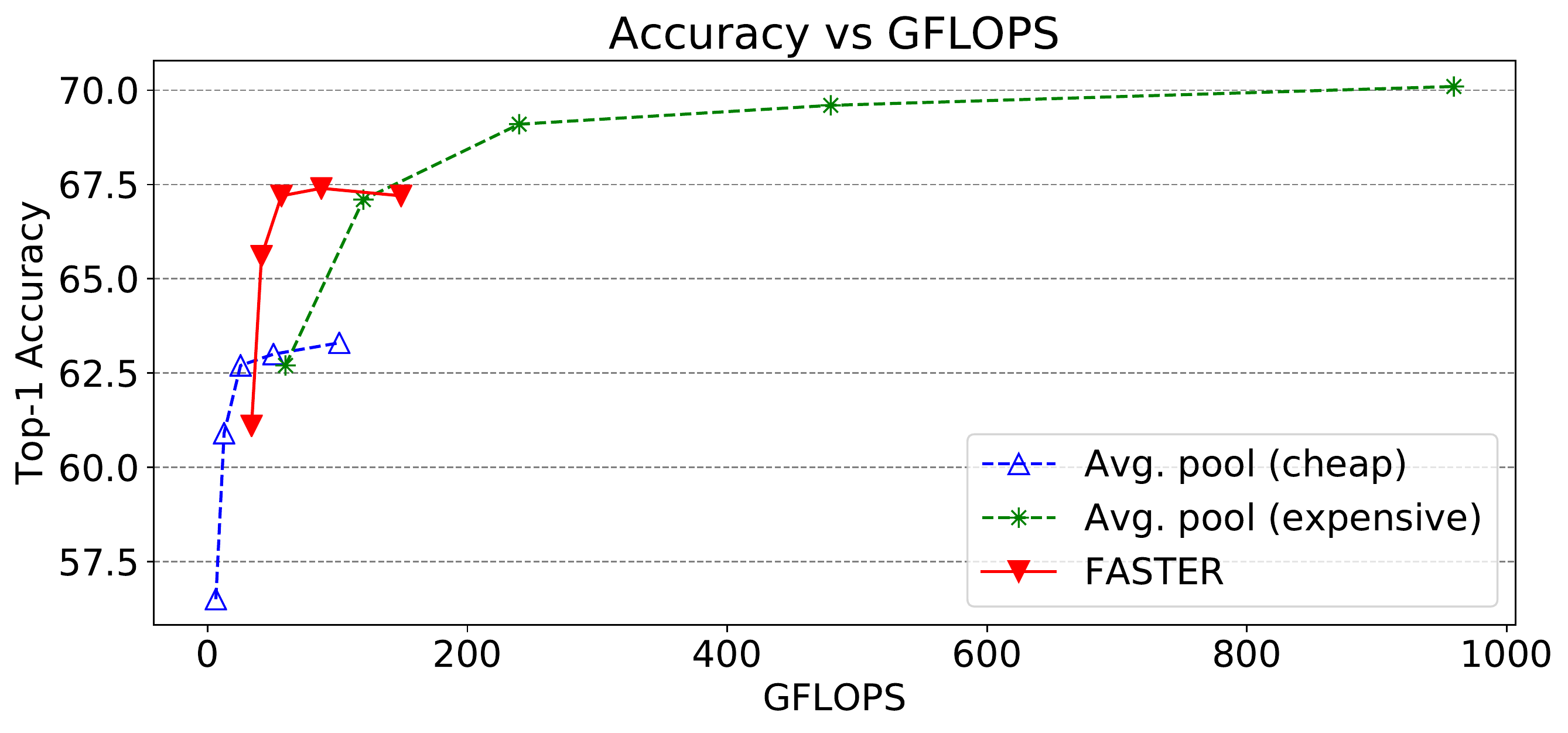}
\caption{
    {\bf \oursfr \vs typical video classification frameworks with average pooling.} We show accuracy \vs GFLOPs of different methods as a function of the number of clips. We compare the results of \ours from Table~\ref{tab:rnn_variations} with average pooling of all expensive (Avg. pool (expensive)) and lightweight models (Avg. pool (cheap)).
}
    \label{fig:flops_comparison}
\end{figure}

``Concat'' outperforms ``Avg. pool'' by 2.1\% when $N=8$, showing the benefits of learning temporal structure. However, as $N$ increases, the performance of ``Concat'' decreases, and by $N=32$, the accuracy is 8.1\% lower than $N=8$. This shows that  learning to aggregate over a longer time period is difficult. Interestingly, ``Concat'', and all recurrent networks have similar performances when the number of clips is small ($N=2,4$), showing that the advantage of RNNs is not evident for short sequences. As the number of clips increases ($N\ge16$), recurrent networks show their strength in modeling long sequences and perform better than the simple ``Concat'' baseline. 

Of all the methods, \ours achieves the best performance for long sequences. For example, \ours outperforms LSTM by 2.5\% and GRU by 1.4\% when $N=32$.
The results show that our proposed \ours is beneficial for long sequence learning, even when sequence samples come from different distributions.

\paragraph{\oursfr \vs. average pooling.} 
How well does the \oursfr framework do with respect to the prevailing average pooling? Recall that our main focus is not only to increase accuracy, but also to reduce computational cost. For this, we compare the results of \ours from Table~\ref{tab:rnn_variations} with the traditional framework (in Fig.~\ref{fig:intro} (a)),  where all clips are cheap (or all clips are expensive) and aggregated with average pooling. These can be considered as the lower and upper bound of the \oursfr framework. We plot the GFLOPs \vs. accuracy comparisons in Fig.~\ref{fig:flops_comparison}.  We use clip length $L=8$ frames, and consider the number of clips $N=\{2, 4, 8, 16, 32\}$, like Table~\ref{tab:rnn_variations}. 
When $N=4$, it takes 119 GFLOPs for ``Avg. pool (expensive)'' to achieve 67.1\%. \oursfr can achieve comparable results with only half the GFLOPs. Comparing with  ``Avg. pool (cheap)'' when $N=16$, \oursfr is over 4\% better with similar amount of GFLOPs.

Note that there is still an accuracy gap between \oursfr and ``Avg. pool (expensive)''.  In the next experiment, we introduce more expensive clips and study the optimal ratio between the  expensive and cheap clips, achieving superior results.

\begin{table}[t]
\resizebox{\columnwidth}{!}{
\centering
\begin{tabular}{c|cc|cc|cc}
    \hline
    \multirow{2}{*}{\#E :\#C} & \multicolumn{2}{c|}{8-frames\x32-clips}     &  \multicolumn{2}{c|}{16-frames\x16-clips} & \multicolumn{2}{c}{32-frames\x8-clips}  \\
   & V$@$1 & GFLOPs &  V$@$1   &  GFLOPs  & V$@$1 &  GFLOPs  \hspace{-1em}  \\
\shline
    All E  &  70.6   &     $982.4$     &   72.9  &     $980.2$    &   74.9    &    $979.2$  \\ 
    1:1   &  70.5   &     $553.6$      &   72.9  &    $552.0$    &      \bf{74.6}    &    $550.4$  \\
    1:3   &  70.1   &     $339.2$      &   72.1   &    $337.6$     &     73.3    &    $336.0$  \\
    1:7   &  69.6   &     $230.4$      &   \bf{71.7}   &   $230.4$     &    71.3     &    $228.8$  \\
    1:15  & 69.0   &     $176.0$       &  69.9   &   $176.0$       &       &              \\
    1:31  & 67.2   &    $150.4$      &       &      &      &   \\ \hline
\end{tabular}
    }
\caption{{\bf Optimal ratio between the  expensive and cheap clips.}
    For a fixed number of total frames, we vary the number of clips $N$ and clip length $L$, and measure accuracy and GFLOPs on Kinetics. Empty cells correspond to combinations that are not feasible. ``E'' denotes the expensive model, while ``C'' denotes the cheap model. 
GFLOPs combine the costs of the cheap, expensive and FAST-GRU models together.
}
\label{table:baseline}
\end{table}

\paragraph{Better accuracy/FLOPs trade-offs.}
\label{sec:mix_feature_combination}
There are three parameters in our framework that affect accuracy and computational cost:  the clip length ($L$), the number of clips ($N$), and the proportion of expensive and lightweight clips used as input pattern (\#E :\#C).  We now provide an empirical study of these parameters to achieve the best trade-off.  Since we are interested in the behavior of \oursfr, we ``fix" the input data by keeping the number of frames constant, which spans the entire video. Thus, $L\times N=256$ for all settings. Given $N$ clips, we experiment with input patterns of $1:x$. It is the ratio between the number of expensive clips and cheap clips, $x \in \{1, 3, 7, 15, 31\}$. For example, when $x=3$ and $L=16$, there will be 4 expensive clips and 12 cheap clips. Additionally, we evaluate the case where all the inputs are expensive clips ($x=0$).

Results are shown in Table~\ref{table:baseline}. As expected, a higher ratio of expensive models leads to a higher accuracy. For example, when $L=8$, there is a large gap of 1.8\% when $x$ goes from $31$ to $15$, with a modest increase of $25.6$ GFLOPs. The gap is less obvious for higher ratios of expensive models.  A remarkable observation is that \oursfr achieves 70.1\% when $x=3$, which is the same accuracy as the average pooling baseline with all expensive clips from Table~\ref{table:backbone}. But \oursfr only consumes 35\% of its GFLOPs. Even more, for $x=1$, the proposed method outperforms the average pooling baseline in Table~\ref{table:backbone}, at 57\% of the cost. This shows the benefit of learning temporal aggregation of hybrid inputs. 

We also observe that for a fixed cost, it is often beneficial to use longer clip length $L$. For example, for a fixed proportion of cheap models $x=3$, the accuracy of  32-frames$\times$8-clips is higher than 16-frames$\times$16-clips, and that in turn is considerably higher than 8-frames$\times$32-clips. In other words, it is better to have fewer clips with more frames. However, for higher ratios of cheap clips (\eg, $x=7$), this pattern does not hold.

\begin{table}[t]
\centering
\resizebox{0.95\columnwidth}{!}{
\begin{tabular}{l|c|c|c}
\hline
 Model   & Time (secs) & GFLOPS   &  Accuracy \\
\shline
 \shortstack{All C (Avg pool) } & 15.1    &   101.3      &     66.7     \\
\hline
    \shortstack{All E (Avg pool)}  & 75.7  & 959.3      &   74.5   \\
\hline
    \shortstack{FASTER32}  & 35.3   & 550.4 & 74.6  \\
\hline
\end{tabular}
}
\caption{{\bf Comparisons of runtime between different methods.} The results of \oursfr is reported in the 32-frames\x 8-clips setting.}
\label{tab:comp_runtime}
\end{table}

\begin{table}[t]
\resizebox{0.95\columnwidth}{!}{
\centering
\begin{tabular}{c|cccc}
    \hline
    \multirow{3}{*}{\#E :\#C} &  \multicolumn{4}{c}{32-frames\x 8-clips  }  \\
                                    & \multicolumn{2}{c}{R(2+1)D-50} & \multicolumn{2}{c}{R3D-50}   \\
                                    & V$@$1  &  GFLOPs &  V$@$1   &  GFLOPs \\
\shline
    All E (Avg. pool)  &  74.5 & 959.3 &  74.4 & 942.7  \\ \hline
    1:1  (\ours)             &  74.6 & 550.4 &  75.3 & 542.1  \\\hline
    1:3  (\ours)             &  73.3 & 336.0  & 73.9 & 331.9  \\ \hline
\end{tabular}
}
\caption{\textbf{Compare different architectures as the expensive clip-level backbone.} Our \oursfr framework can generalize to different architectures, and R3D even performs slightly better than R(2+1)D.} 
\label{table:comp_backbone}
\end{table}

We choose the parameter setting $L=16$ and $x=7$ as the best inexpensive configuration, and call this setting FASTER16 (\ie, $L=16, x=7$). It indicates that it is possible to achieve better performance with shorter clip length $L$, which reflects the importance of learning a good aggregation function. When $L=32$ and $x=1$, the proposed framework achieves 74.6\% accuracy with 550.4 GFLOPs. We denote this setting as FASTER32 (\ie, $L=32, x=1$) and it is the best model when the cost is around $\sim$550 GFLOPs .

\oursfr also works when all input features come from the same network. When the inputs are all expensive features, \oursfr outperforms the average pooling baseline from Table~\ref{table:backbone} in all settings, with a negligible increase of FLOPs from \ours.

\paragraph{Runtime \vs FLOPs.}
We measure the runtime speed of different methods on a TITAN X GPU with Intel i7 CPU. We sum up the runtime over 100 Kinetics videos. The results are listed in Table~\ref{tab:comp_runtime}, which confirms that the theoretical FLOPs are consistent with the runtime for our experiments. The reduction of FLOPs is roughly proportional to the decrease of runtime.

\paragraph{Generalize to different backbones.}
Our \oursfr framework does not make any assumption about the clip-level backbones. To demonstrate that \oursfr can generalize to different backbones, we replace the expensive model, \ie, R(2+1)D-50, with R3D-50~\cite{tran2018closer} and keep everything else unchanged. In Table~\ref{table:comp_backbone}, R3D even works slightly better than R(2+1)D. Comparing FASTER32 with the average pooling baseline,  using R3D leads to a larger improvement of  0.9\%.

\begin{table}[t]
\begin{center}
\resizebox{\columnwidth}{!}{
    \begin{tabular}{l|c|c|c}
\hline
        Model                                                    &  Top-1 & \shortstack{ImageNet \\ pre-train} & GFLOPs$\times$clips   \\
    \shline
    I3D~\cite{carreira2017quo}                          &  72.1    & \checkmark      &   108$\times$4      \\
\hline
    S3D~\cite{Xie_2018_ECCV}                            &  72.2    & \checkmark     &    66.4$\times$N/A    \\
\hline
    MF-Net~\cite{chen2018multi}                         &  72.8    & \checkmark           &     11.1$\times$50          \\ 
\hline
    A$^2$-Net~\cite{chen20182}                                             &    74.6       &  \checkmark                        &          40.8$\times$30     \\
\hline
    S3D-G~\cite{Xie_2018_ECCV}                            &  74.7   & \checkmark        &    71.4$\times$N/A    \\
\hline
    NL I3D-50~\cite{wang2018non}                                     &  76.5     & \checkmark        &      282$\times$30     \\
\hline
    NL I3D-101~\cite{wang2018non}                                     &  77.7   & \checkmark         &      359$\times$30     \\
\hline
\hline
    I3D~\cite{carreira2017quo}                &  68.4        &  -   &   108$\times$4 \\
\hline
    STC~\cite{diba2018spatio}                &  68.7        &  -   &   N/A$\times$N/A \\
\hline
    ARTNet~\cite{wang2017appearance}              &  69.2        &  -   &    23.5$\times$250    \\
\hline
    S3D~\cite{Xie_2018_ECCV}                &  69.4        &   -  &   66.4$\times$N/A \\
\hline
    ECO~\cite{zolfaghari2018eco}            & 70.0          & - & N/A$\times$N/A \\
\hline
    R(2+1)D-34~\cite{tran2018closer}                    &  72.0        &  -   &   152$\times$115      \\
\hline\hline
    FASTER16                & 71.7   &    -      &  14.4$\times$16   \\
\hline
    FASTER32               & {\bf 75.3}   &    -      & 67.7$\times$8     \\
\hline

    \end{tabular}}
\end{center}
    \caption{{\bf Comparisons to the \sota methods on Kinetics}.
   The inputs for all the methods are RGB frames. For ``GFLOPs $\times$ clips'', we report the cost of a single clip and the number of clips used. ``N/A''' indicates the number of clips are not reported in the paper.
    }
\label{expr:final_results_K}
\end{table}
\begin{table}[t]
\begin{center}
\resizebox{\columnwidth}{!}{
\begin{tabular}{l|cc|c|c}
\hline
    \multirow{2}{*}{Model} &     \multicolumn{2}{c|}{Pre-train} & \multirow{2}{*}{UCF-101}  & \multirow{2}{*}{HMDB-51}  \\
                                  &         I & K  &  &  \\
\shline
    STC~\cite{diba2018spatio}              & - & \checkmark &   95.8      &   72.6      \\
\hline
    ARTNet~\cite{wang2017appearance}       & - & \checkmark &   94.3      &   70.9   \\
\hline
    ECO~\cite{zolfaghari2018eco}           & - & \checkmark  &   93.6      &    68.4 \\
\hline
    R(2+1)D-34~\cite{tran2018closer}       &  - & \checkmark  &     96.8    &    74.5   \\
\hline
    I3D~\cite{carreira2017quo}             & \checkmark  & \checkmark &     95.6    &   74.8  \\
\hline
    S3D~\cite{Xie_2018_ECCV}               & \checkmark  & \checkmark   &     96.8     &    75.9 \\
\hline\hline
    FASTER32                & - & \checkmark    &    96.9      &   75.7   \\
\hline
\end{tabular}}
\end{center}
    \caption{{\bf Comparisons on UCF-101 and HMDB-51.} ``I'' denotes pre-training on ImageNet, while ``K'' denotes pre-training on Kinetics.}
\label{expr:ucf_101}
\end{table}
\begin{figure}[t]
\centering
\includegraphics[width=0.95\linewidth]{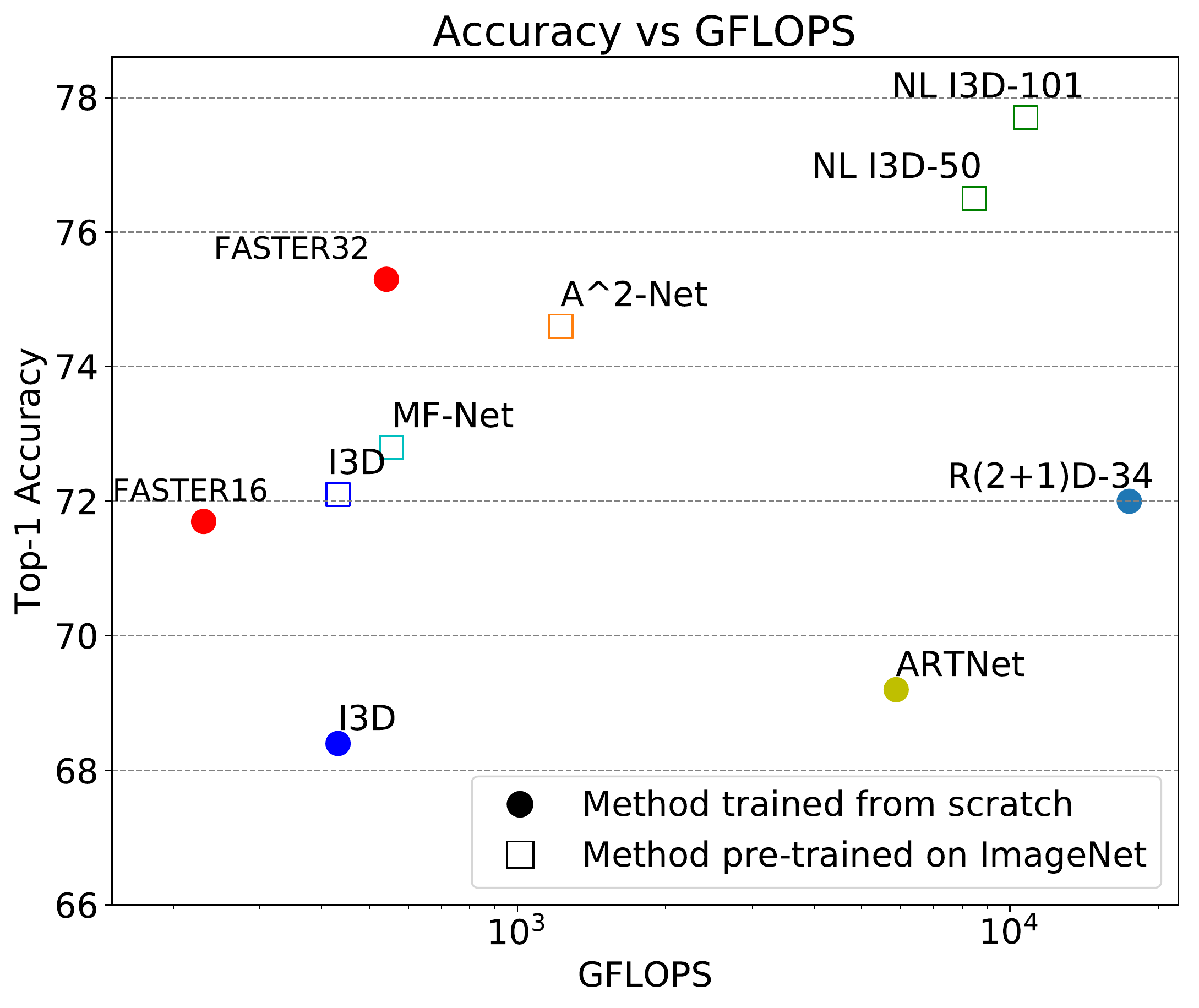}
    \caption{{\bf Log-scale GFLOPs  \vs accuracy  comparisons  on Kinetics.} In this chart, optimal methods are closer to the top-left corner. The two proposed variants of \oursfr are the closest to the corner.
    }
\label{fig:final_cmp}
\end{figure}

\paragraph{Comparison to \sota}
We now compare the \oursfr framework to \sota methods across three most popular video  datasets, \ie, Kinetics, UCF-101 and HMDB-51. We only include methods that use RGB frames as inputs. Results on Kinetics are shown in Table~\ref{expr:final_results_K}. These same numbers are also visualized in terms of accuracy \vs GFLOPs in Fig.~\ref{fig:final_cmp}. Our proposed \oursfr framework achieves the best accuracy/GFLOPs trade-offs.  This is specially impressive since  \oursfr is not pre-trained on ImageNet, and the clip-level backbones used are not the most cost effective ones. In this comparison, we use FASTER32 with R3D as its backbone. FASTER32 outperforms R(2+1)D-34 by 3.3\% with only 4\% of its GFLOPs. FASTER16 outperforms I3D trained from scratch by 3.3\% (71.7\% \vs 68.4\%), using only half of I3D's cost (432 GFLOPs).  Even though we do not leverage pre-training, our framework outperforms all existing pre-trained models except NL-I3D. However, FASTER32 only uses 5\% of the GFLOPs of NL-I3D. Our FASTER32 model outperforms $A^2$-Net by 0.7\%, while the cost is reduced by over 50\%. The results validate the effectiveness of our \oursfr framework, showing it is promising to leverage the combination of expensive and lightweight models for better accuracy/FLOPs trade-off.

Competitive results are also achieved on UCF-101 and HMDB-51 (Table~\ref{expr:ucf_101}). It shows learning the aggregation function on larger datasets can facilitate the generalization on smaller datasets. The improvements are less profound as the performance on small datasets tends to be saturated.
\section{Conclusion}
\label{sec:conclusion}
In this paper, we propose a novel framework called \oursfr for efficient video classification. To exploit the spatio-temporal redundancy of the video sequence, we combine the representations from both expensive and lightweight models.  We propose a recurrent unit called \ours to learn to  aggregate mixed representations. Experimental results show that our \oursfr framework has significantly better accuracy/FLOPs trade-offs, achieving the state-of-the-art accuracy with $10\times$ less FLOPs.

{\small
\bibliographystyle{ieee}
\bibliography{egbib}
}

\end{document}